# A Novel Semi-supervised Data-driven Method for Chiller Fault Diagnosis with Unlabeled Data

Bingxu Li[a,b], Fanyong Cheng[a,c], Xin Zhang[d*], Can Cui[e], Wenjian Cai[a]

[a.] *School of Electrical and Electronic Engineering, Nanyang Technological University, 50 Nanyang Avenue, 639798, Singapore*
[b.] *Energy Research Institute @ NTU (ERI@N), Interdisciplinary Graduate Programme, Nanyang Technological University, Singapore*
[c.] *School of Electrical Engineering, Anhui Polytechnic University, Wuhu, 241000, China*
[d.] *College of Electrical Engineering, Zhejiang University, Hangzhou, 310027, China*
[e.] *Building and Construction Authority (BCA)，608550, Singapore*

## Abstract

In practical chiller systems, applying efficient fault diagnosis techniques can significantly reduce energy consumption and improve energy efficiency of buildings. The success of the existing methods for fault diagnosis of chillers relies on the condition that sufficient labeled data are available for training. However, label acquisition is laborious and costly in practice. Usually, the number of labeled data is limited and most data available are unlabeled. The existing methods cannot exploit the information contained in unlabeled data, which significantly limits the improvement of fault diagnosis performance in chiller systems. To make effective use of unlabeled data to further improve fault diagnosis performance and reduce the dependency on labeled data, we proposed a novel semi-supervised data-driven fault diagnosis method for chiller systems based on the semi-generative adversarial network, which incorporates both unlabeled and labeled data into learning process. The semi-generative adversarial network can learn the information of data distribution from unlabeled data and this information can help to significantly improve the diagnostic performance. Experimental results demonstrate the effectiveness of the proposed method. Under the scenario that there are only 80 labeled samples and 16000 unlabeled samples, the proposed method can improve the diagnostic accuracy to 84%, while the supervised baseline methods only reach the accuracy of 65% at most. Besides, the minimal required number of labeled samples can be reduced by about 60% with the proposed method when there are enough unlabeled samples.

---

[*]Corresponding author. *Email address*: zhangxin_ieee@zju.edu.cn

**Nomenclature**

*Abbreviations*

| | | | |
|---|---|---|---|
| cf | condenser fouling | GAN | generative adversarial network |
| COP | coefficient of performance | HVAC | heating, ventilation, and air conditioning system |
| D | discriminator | nc | non-condensable gas in refrigerant |
| eo | excess oil | NN | neural network |
| FD | fault diagnosis | PCA | principle component analysis |
| fwc | reduced condenser water flow | rl | refrigerant leak |
| fwe | reduced evaporator water flow | ro | refrigerant overcharge |
| G | generator | SVM | support vector machine |

*Symbols*

| | | | |
|---|---|---|---|
| $b_m$ | size of the minibatch | $p_D(y=c_i \vert x)$ | the probability that the sample $x$ is assigned to the $i^{th}$ class by the discriminator |
| $c_i$ | the label of the $i^{th}$ class | $t^{(i)}$ | the actual label of the $i^{th}$ labeled data sample |
| $D(\cdot)$ | the function of discriminator | $x^{(i)}$ | the $i^{th}$ data sample |
| $G(\cdot)$ | the function of generator | $x_l^{(i)}$ | the $i^{th}$ labeled data sample |
| $l_i$ | the $i^{th}$ output of the discriminator network | $x_u^{(i)}$ | the $i^{th}$ unlabeled data sample |
| $L$ | loss function | $\tilde{x}^{(i)}$ | the $i^{th}$ generated data sample |
| $m$ | the number of samples | $y$ | the predicted label |
| $N$ | the number of classes | $z^{(i)}$ | the $i^{th}$ random noise sample input to the generator |
| $n_l$ | the number of labeled samples | $\theta_d$ | parameters of the discriminator |
| $n_u$ | the number of unlabeled samples | $\theta_g$ | parameters of the generator |

# 1. Introduction

*1.1 Background*

Heating, ventilation, and air conditioning (HVAC) systems are responsible for a large portion of energy consumption in the commercial buildings [1]. In U.S. commercial buildings, about 33% of the energy consumption can be attributed to HVAC systems [2]. In Singapore, which is a tropical country with hot and humid weather throughout the year, the proportion can be as high as 60% [3]. Improving the energy efficiency of HVAC system is an essential goal in facility management industry [4]. Among the components in HVAC systems, the chiller is the major energy consumer, especially in the tropics. Chiller faults will lead to uncomfortable indoor environment and serious energy waste. For example, the "soft faults" [5], such as a slow loss of refrigerant or the condenser fouling, are difficult to detect and may remain unnoticed for a long time which will reduce the chiller's coefficient of performance (COP) and thus lead to excessive energy waste. Thus, applying the accurate fault diagnosis (FD) techniques to the practical chiller systems can significantly reduce the energy consumption and improve the energy efficiency of the buildings [6].

*1.2 Review of the existing data-driven fault diagnosis methods for chillers*

Several studies have focused on the development of the chiller FD methods in recent years. The existing chiller FD methods can be classified into two main categories: model-based methods and the data-driven methods. The model-based methods [7] rely on the detailed physical model of the chiller system. However, the chiller systems are often complex systems with various kinds of controllers, devices, and sensors. The complexity of the chiller systems limits the applicability of the model-based methods.

Compared with the conventional model-based methods, the data-driven methods do not require to develop complex chiller models. The deep understanding of the physical principles in chiller system is also no longer needed. The data-driven methods can learn the fault patterns from the historical information by the machine learning techniques and then predict the faults reliably. With the development of the building management systems (BMS), many historical monitoring data can be collected and utilized. Besides, the machine learning theories and algorithms are becoming mature, with plenty of documented information and tools available. Due to the above tendencies, the data-driven FD methods have attracted more and more attention and become dominant in the field of the fault diagnosis for not only the chiller systems but also other HVAC subsystems recently. The existing data-driven FD methods can be classified into two categories: the supervised methods and the semi-supervised methods.

The supervised methods use the labeled data samples to train the fault classifier. In 2009, Du *et al* developed a FD method for the sensors in variable air volume systems based on wavelet neural network [8]. In 2012, Najafi *et al* developed a diagnostic algorithm for air handling units by using supervised learning techniques [9]. This algorithm is less dependent on model accuracy and more flexible with respect to measurement constraints. In 2013, a fault detection method is proposed in [10] based on the support vector data description (SVDD) for the chiller system. This method can achieve good detection performance and can properly addresses nonlinear, non-Gaussian and wide-range process variables. In [11], Zhao *et al* designed a novel framework for intelligent fault diagnosis of chiller based on the three-layer Bayesian network. This method can well deal with the incomplete information. In [12], Zhao *et al* proposed a new FD strategy for chiller systems, combining the exponentially-weighted moving average (EWMA) control charts and support vector regression. The ratios of correctly diagnosed faults can be significantly improved. In 2014, Yan *et al* [13] designed a hybrid FD algorithm incorporating auto-regressive model with exogenous variables and support vector machines. This method can achieve lower false alarm rates and higher prediction accuracy. In [14], Zhao *et al* considered the chiller FD problem as the one-class classification problem and used SVDD algorithm to diagnose different faults. In [15], Du *et al* integrated the basic neural network and auxiliary neural network to detect the abnormities in air handling units. The clustering analysis is utilized to classify faulty conditions in the buildings. In 2016, a PCA-R-SVDD-based fault detection method is proposed in [16] for the chiller systems. The SVDD model is developed in the residual subspace which is determined by the principle component analysis (PCA). This method is sensitive to six common chiller faults. In [17], Yan *et al* proposed a FD strategy for air handling units based on decision tree. The FD results of this strategy are interpretable. In [18], Li *et al* proposed a two-stage data-driven FD method for building chillers based on the linear discriminant analysis (LDA). This method can successfully diagnose seven typical faults and report unknown faults. The severity level of the fault can also be determined. In [19], He *et al* proposed a FD method for chillers using Bayesian network classifier with probabilistic boundary. After integrating the probabilistic boundary, the false alarm rate can be significantly reduced. In 2017, Wang *et al* [20] merged the multi-source information (MI) and the distance rejection (DR) into the Bayesian network (BN) to identify the possible faults in chiller systems. The FD performance can be significantly improved, and new types of faults can be identified accurately. In [21], Yan *et al* developed a novel hybrid method to detect faults in chiller systems by combining extended Kalman filter and recursive one-class support vector machine. The faulty data is not required in the training phase and typical chiller faults can be detected accurately. In 2018, Wang *et al* [22] combined the BN and PCA to diagnose chiller faults. The FD accuracy can be improved especially for faults at slight levels. In 2019, a practical chiller FD method is developed by Wang *et al* [23] by introducing discretization to BN. The robustness of the FD is improved. In [24], Han *et al* proposed a FD method based on least squares support vector machine (LS-SVM). The LS-SVM is optimized by cross validation to obtain better FD performance. In [25], Fan *et al* designed a FD method based on PCA-SMOTE (synthetic minority oversampling technology)-SVM. This method can transfer the prior knowledge of the centrifugal chiller to the screw chiller effectively. In 2020, Zhang *et al* [26] applied ensemble learning which integrates several machine learning models to diagnose faults in the refrigeration system. The integrated model can obtain high accuracy on both local faults and system faults.

All the above-mentioned FD methods are based on the supervised machine learning algorithms relying on enough training data samples. Recently, some researchers started to apply the semi-supervised learning methods to this field. Semi-supervised learning is a kind of approach which incorporates both supervised learning technique and unsupervised learning technique. In [27], Beghi *et al* proposed a semi-supervised one-class SVM classifier used as the novelty detection system to identify possible faults. In [28], Beghi *et al* proposed a semi-supervised FD approach which makes no use of the priori knowledge about abnormal phenomena. The PCA model and a reconstruction-based contribution method are used to detect anomalies. Then, a decision table is used to diagnose the specific faults. In 2018, Yan *et al* proposed a semi-supervised learning FD framework for air handling units [29]. With this method, the number of training samples can be enhanced by inserting confidently labeled samples from the unlabeled set. This FD method can achieve good performance when only a small number of faulty training data samples are available. In 2020, Yan *et al* [30] proposed a chiller FD method integrating an unsupervised learning framework called generative adversarial network (GAN). The artificial fault training samples can be generated to diversify and re-balance the training dataset. The diagnostic accuracy can be improved obviously especially for the case when the number of the faulty samples is insufficient. Another work by Yan *et al* [31] utilized GAN to address the imbalanced data problem in fault diagnosis for air handling units. Compared with the previous method, the method in [31] can reduce the dependency on the labeled data of faulty conditions.

Although the above works have made great progress in the field of chiller FD, there still exists some problems which are not well addressed:
- The existing supervised methods rely on enough labeled training data samples. However, the collected dataset from practical industrial processes or the BMS are commonly partially labeled. In most of the time, the available labeled data samples are usually not enough to train an accurate classifier. Although the labeled data samples can also be obtained by conducting the experiments in which the faults are manually introduced, acquiring data samples through these experiments are time-consuming and laborious.
- The existing semi-supervised methods can well address the data imbalanced problem that the number of fault samples is insufficient. These methods can reduce the requirement for the fault data samples. However, these methods still rely on the condition that there are sufficient normal data samples available for training. In practice, it is very common that plenty of data samples are completely unlabeled, which means we do not know these data correspond to normal or faulty working conditions. In such a scenario, a semi-supervised FD method should be developed which can utilize the unlabeled data to improve the FD performance.

*1.3 Contributions*

To address the abovementioned research problems, we proposed a novel semi-supervised data-driven fault diagnosis method for chiller systems based on the semi-generative adversarial network. Both the labeled data and the unlabeled data can be used to train the classifier. The information contained in the large number of unlabeled data can be efficiently exploited and the FD performance can be further improved. The main contributions of the proposed FD method are listed as follows:

1) **A novel chiller FD framework which can efficiently utilize unlabeled data.** For the existing FD methods, much information contained in the large number of unlabeled data cannot be exploited and this loss of information greatly limits the improvement of the FD performance in chiller systems. The proposed chiller FD framework incorporates both unlabeled and labeled samples into learning process. It can learn the information about the distribution of data from unlabeled samples. This information can help to significantly improve the FD performance. The proposed method fills the current research gap in the FD field of HVAC on how to utilize the information contained in the massive unlabeled data.

2) **Less dependency on the labeled data.** With the information obtained from unlabeled data, much less labeled data are required for obtaining the desired accurate classification model. The proposed method demonstrates its superiority over other existing FD methods especially in the practical scenario with limited number of labeled samples and plenty of unlabeled samples.

3) **A comprehensive comparative study with different number of labeled samples and unlabeled samples.** Experiments were conducted by varying the number of labeled and unlabeled samples in the training set. The minimal required labeled data size for different desired diagnostic accuracy using the proposed method is studied.

The schematic overview of the proposed fault diagnosis framework for chillers is shown in Fig. 1. The outliers in the original dataset is removed in the data preprocessing step. Then, both unlabeled and labeled samples are used to train the classifier. After the training, the obtained discriminator network is used as the fault classification model in the online application. The classification model takes in the real-time operating data and then outputs the corresponding fault type.

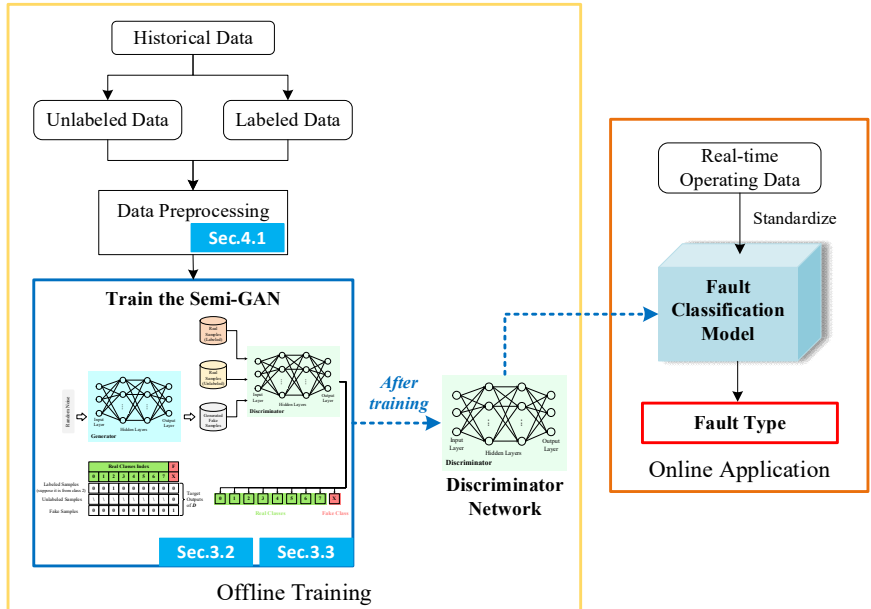

Fig. 1. Schematic overview of the proposed fault diagnosis framework for chiller systems

*1.4 Organization of the paper*

The rest of the paper is organized as follows: Section 2 describes the details about chiller systems, common faults in chillers and the experiment data used in this paper. Section 3 presents the principles of the semi-GAN. The improving techniques for training semi-GAN are also introduced. Section 4 describes the proposed FD framework for chiller systems. The results and discussions are presented in Section 5. Finally, this paper is concluded in Section 6.

## 2. Description of the chiller system

*2.1 Basic Working Principle of the Chiller System*

Generally, a chiller system consists of 4 main components: the evaporator, the condenser, the expansion valve and the compressor. Fig. 2 shows the simple schematic of a chiller system with 4 main components. The refrigeration cycle in chillers can be illustrated as follows: After taking away the heat of the buildings, the chilled water enters in the evaporator and transfers the heat to the refrigerant with lower temperature. The refrigerant evaporates and the refrigerant vapor enters in the compressor where the vapor with low pressure and low temperature is compressed into the vapor with high pressure and high temperature. Flowing out of the compressor, the refrigerant vapor enters in the condenser and rejects the heat to the cooling water. In the condenser, the refrigerant vapor condenses into the liquid. The liquid refrigerant with high pressure and high temperature is expanded into the liquid with low pressure and low temperature through the expansion valve. Then, the refrigeration cycle repeats.

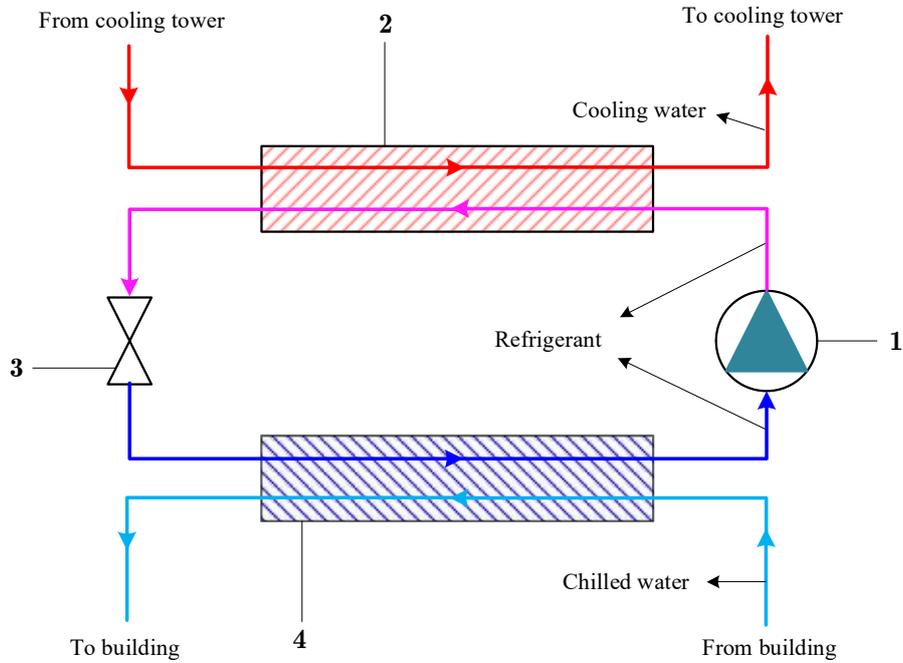

Fig. 2.  Schematic diagram of a chiller system with 4 main components:
1—Compressor, 2—Condenser, 3—Expansion Valve, 4—Evaporator

*2.2 Experimental Data of a Practical Chiller System*

The chiller experimental data used in this study was collected by ASHRAE Project 1043-rp [32]. The chiller system studied is a McQuay PEH048J 90-t (316 kW) centrifugal chiller. It consists of a centrifugal compressor, a shell-and-tube evaporator, a pilot-driven expansion valve, and a shell-and-tube condenser.

We study the following 7 typical chiller faults which are the most harmful ones considering the occurrence frequency and the economic losses [32]: reduced condenser water flow (*fwc*), non-condensable gas in refrigerant (*nc*), condenser fouling (*cf*), reduced evaporator water flow (fwe), excess oil (*eo*), refrigerant overcharge (*ro*), and refrigerant leak (*rl*). The experimental data under these faulty conditions were collected.

For each faulty condition, 4 different levels of severity are considered. All these faults are manually introduced to the chiller system. For instance, the '*fwc*' fault at severity 1 is introduced by reducing the condenser water flow by 10% while the '*fwc*' fault at severity 4 is introduced by reducing the condenser water flow by 40%. Other details about how to introduce these chiller faults with different severity levels can be found in [32].

For the simplicity, in this study, different faults are assigned with different fault labels, as shown in Table 1:

Table 1: Fault labels for 7 typical chiller faults

| Fault Label | 0 | 1 | 2 | 3 | 4 | 5 | 6 | 7 |
|---|---|---|---|---|---|---|---|---|
| Fault Type | *normal* | *cf* | *eo* | *fwc* | *fwe* | *nc* | *rl* | *ro* |

The experimental data contains 65 features, such as condenser in/out temperature, evaporator water flow rate, condenser water flow rate and so on. The data of each feature was sampled at 10 second intervals and includes transient data between the steady state operating conditions [32]. Out of 65 features, the value of 4 features is constant under different faults so these 4 features are useless for the fault classification. Only 61 features are used in this study.

For each severity level of each fault, the data were collected for one day under 27 predefined operating conditions, containing about 5000 samples. For the normal state without faults, dataset was collected for several days, containing more than 30000 samples. 27 operating conditions were obtained by varying three control variables: the chiller cooling load, the condenser water entering temperature, and the chilled water temperature. Using these three variables each at three levels resulted in 27 different operating conditions [32].

# 3. Semi-Generative Adversarial Network

## 3.1 Generative adversarial network

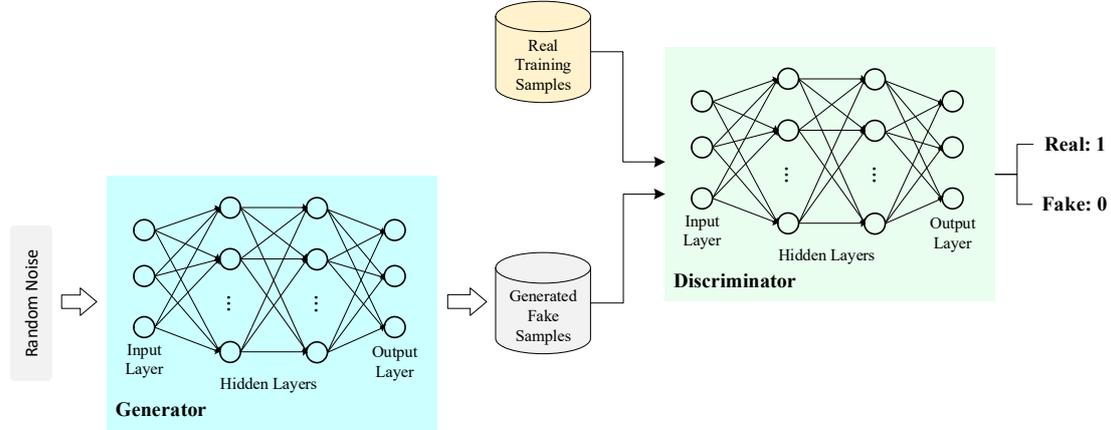

Fig. 3. Schematic diagram of the generative adversarial network (GAN)

The generative adversarial network (GAN) was firstly introduced by Goodfellow *et al* in [33]. GAN has been widely used in the field of computer vision. The typical applications of GAN include generating images, natural image synthesis, improving image compressions, face recognition, etc. In the field of chiller fault diagnosis, Yan *et al* [30] used GAN to generate synthetic faulty training samples. The number of faulty training samples in the training set are increased and the imbalanced data problem that the fault samples are insufficient can be well addressed.

The generative adversarial network framework comprises two neural networks, the generator *G*, and the discriminator *D*. The schematic of a typical GAN framework is shown in Fig. 3. The generator *G* takes a random noise vector as the input and outputs a fake data sample. Then, fake data samples together with real data samples are input into the discriminator *D*. *D* outputs a prediction label about whether this data sample is real or fake. The training objectives of *G* and *D* are conflicting with each other: the objective of *D* is to discriminate the fake samples from real-world samples as accurately as possible while *G* is trained to generate data samples as close as possible to the real-world samples to "fool" the discriminator. The training process for the GAN can be described as follows. Let $\{x^{(1)}, x^{(2)}, ..., x^{(m)}\}$ be the real-world data samples and $\{z^{(1)}, z^{(2)}, ..., z^{(m)}\}$ be the random noise samples input to the generator. $G(\bullet)$ and $D(\bullet)$ denote the function of generator and discriminator, respectively. $\theta_d$ and $\theta_g$ are the parameters of *D* and *G*, respectively. In each training step,

- Firstly, *G* produces data samples from random noise vectors. These generated fake samples, together with some real-world samples, are presented to *D*. The task of *D* is to learn to classify them as fake (label: 0) or real (label: 1). That is, the parameters of the *D* are updated to minimize the probability that *D* makes a mistake. The updating rule is given by:

$$L_1 = -\frac{1}{m}\sum_{i=1}^{m}\left[\log D\left(x^{(i)}\right) + \log\left(1 - D\left(G\left(z^{(i)}\right)\right)\right)\right]$$
$$\theta_d = \theta_d - \eta\nabla_{\theta_d}L_1$$
(1)

- Secondly, *G* learns to produce data samples as close as possible to the real-world samples. Taking the random noise vectors as input, the parameters of *G* are updated to maximize the probability that *D* makes a mistake, i.e., *D* classifies the generated sample as the real one. It should be noted that the parameters of *D* are fixed when updating parameters of *G*. The updating rule is given by:

$$L_2 = -\frac{1}{m}\sum_{i=1}^{m}\log\left(D\left(G\left(z^{(i)}\right)\right)\right)$$
$$\theta_g = \theta_g - \eta\nabla_{\theta_g}L_2$$
(2)

This process is repeated iteratively. In the end, the generator can generate data samples which can fool the discriminator. The training process of GAN resembles a process of producing fake currency [33]: The generator is analogous to the counterfeiters while the discriminator is analogous to the police. The counterfeiters try to make fake currency to fool the police while the police try to detect all the fake currency made by counterfeiters. The competition between the police and the counterfeiters drives both to improve their methods until the fake currency cannot be distinguished from the genuine currency.

*3.2 Semi-Generative Adversarial Network*

The generative adversarial network is an unsupervised framework. The goal of the conventional GAN is to generate synthetic data samples whose distribution is the same as that of the samples in training sets. This architecture can be extended to the semi-supervised learning framework [34]. Semi-supervised learning is a kind of machine learning method which can utilize both labeled and unlabeled data to train the classifier. Similar with conventional GAN, semi-GAN consists of a generator network and a discriminator network. The main difference is that the discriminator should not only predict the label about whether this data sample is real or fake, but also predict which class the sample belongs to. That is, the discriminator can be regarded as a classifier. Consider a classification problem that has *N* classes. In this case, the discriminator is a classifier with *N*+1 classes. The additional class is the one which indicates whether the sample is real or fake. The sample is assigned to this additional class *N*+1 if this sample is fake. This additional class is called "fake class" in the following part. The structure of the semi-GAN is shown in Fig. 4.

Three types of data are utilized to train the semi-GAN:

- Labeled training data sample: the discriminator (i.e., the classifier) should not only assign the labeled sample to the correct class but also assign this sample to "real". The target output corresponding to its class should be 1 while the target output corresponding to the fake class should be 0.

- Unlabeled training data sample: the discriminator only needs to assign this sample to "real". The target output corresponding to the fake class should be 0. The discriminator does not need to assign this sample to the correct class.

- Generated data sample: the discriminator should assign this sample to "fake", i.e., the target output corresponding to the fake class should be 1 while all the other outputs should be 0.

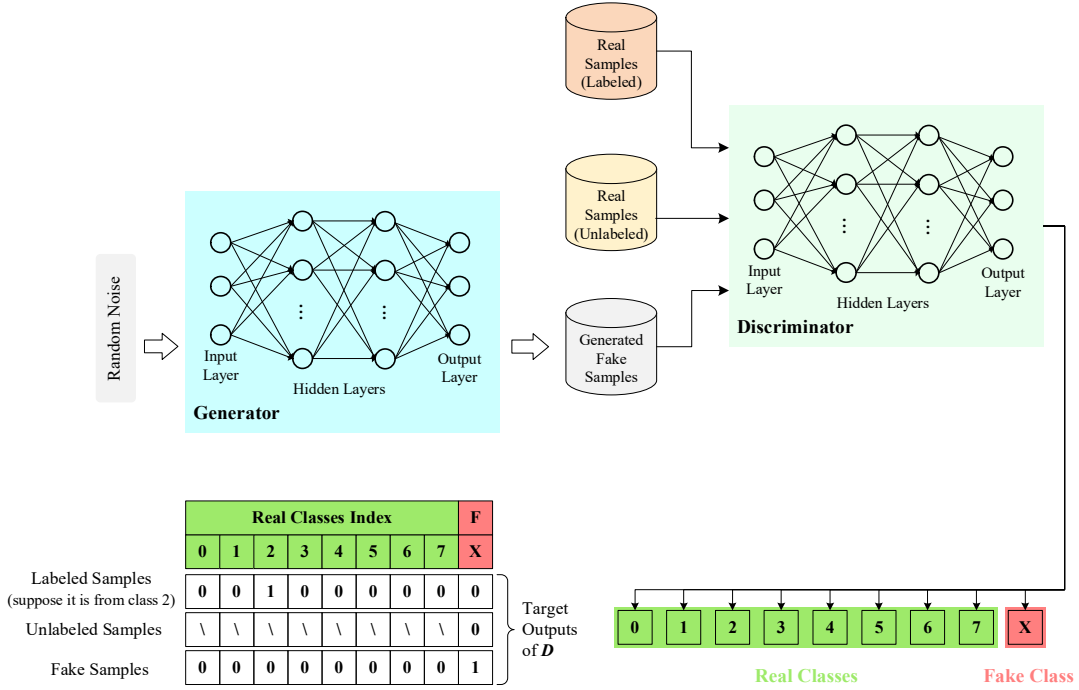

Fig. 4. Schematic diagram of the semi-generative adversarial network (Semi-GAN)

1) Training of the discriminator

Training of the discriminator in the semi-GAN is different from the conventional GAN. Since the roles of different types of data in the training process are totally different, the loss functions for different types of data are different. In the semi-GAN framework, the discriminator is a multi-class classifier. We may then use $p_D(y = c_i | x)$ to denote the probability that the sample $x$ is assigned to the $i^{th}$ class by the discriminator. $y$ is the predicted label and $c_i$ is the label of the $i^{th}$ class (in this study, $c_i = i - 1$). For example, the label of the 1st class is 0 and the label of the 8th class is 7. Generally, the probability $p_D(y = c_i | x)$ is calculated by the softmax function:

$$p_D(y = c_i | x) = \exp(l_i) / \sum_{k=1}^{N} \exp(l_k) \qquad (3)$$

where $N$ is the number of classes (in this study, $N = 8$). $\{l_1, l_2, ..., l_N\}$ are the first $N$ elements in the output vector of the discriminator network.

For the labeled training data samples, we should update the parameters of $D$ to maximize the probability that the data sample is assigned to its corresponding class. This is equivalent to minimizing the following loss function:

$$L_{labeled} = -\frac{1}{m_l} \sum_{i=1}^{m_l} \log\left(p_D(y = t^{(i)} | x_l^{(i)})\right) \tag{4}$$

where $x_l^{(i)}$ is the $i^{th}$ labeled data sample, and $t^{(i)}$ is the actual label of the $i^{th}$ labeled data sample ($t^{(i)} \in [0, 1, 2, ..., N-1]$).

For the unlabeled training data samples, we should update the parameters of *D* to maximize the probability that the data sample is assigned to "real", i.e., not assigned to the fake class. This is equivalent to minimizing the following loss function:

$$L_{unlabeled} = -\frac{1}{m_u} \sum_{i=1}^{m_u} \log\left(1 - p_D(y = N | x_u^{(i)})\right) \tag{5}$$

For the generated training data samples, the parameters of *D* should be updated to maximize the probability that the data sample is assigned to the fake class. This is equivalent to minimizing the following loss function:

$$L_{fake} = -\frac{1}{m_f} \sum_{i=1}^{m_f} \log\left(p_D(y = N | G(z^{(i)}))\right) \tag{6}$$

Then, the parameters of discriminator can be updated by gradient descent according to the corresponding loss function:

$$\theta_d = \theta_d - \eta \nabla_{\theta_d} L_i \quad (i = labeled, unlabeled, fake) \tag{7}$$

2) Training of the generator

Training for the generator in semi-GAN is similar with that in the conventional GAN. Its goal is to generate data samples to fool the discriminator so that the discriminator will regard these samples as "real" ones. That is, the parameters of G should be updated so that the probability that the sample is assigned to "real" can be maximized. It should be noted that the generator does not care which of *N* classes the generated sample should belong to. The updating rule for the parameters of generator is given by:

$$\begin{aligned} L_3 &= -\frac{1}{m_f} \sum_{i=1}^{m_f} \log\left(1 - p_D(y = N | G(z^{(i)}))\right) \\ \theta_g &= \theta_g - \eta \nabla_{\theta_g} L_3 \end{aligned} \tag{8}$$

Training of the discriminator and the generator is implemented alternatively through a series of iterations. The combination of different sources of data samples makes the classifier able to learn from a broader perspective. By learning the pattern and distribution present in unlabeled data, the performance of the classifier can be significantly improved. The obtained model can perform classification much more accurately than it would be if only using very few labeled training samples.

*3.3 Improving technique for training Semi-GAN*

In Semi-GAN, the classifier with *N*+1 outputs is actually over-parameterized and a more efficient training technique can be developed for semi-GAN [34]. According to [34], this technique can significantly improve the training performance for semi-GAN. We briefly introduce this technique as follows:

Consider the discriminator with *N*+1 outputs introduced in 3.2. The output vector of the discriminator is $\{l_1, l_2, ..., l_N, l_{N+1}\}$. Since subtracting a general function from each element in the output vector does not change the output value of the softmax function (3), we can set:

$$l_j \leftarrow l_j - l_{N+1}, \quad j = 1, 2, ..., N \tag{9}$$

This makes $l_{N+1}$ equal to zero all the time. Then, the probability $p_D(y = c_i | x)$ should be calculated by eq.(10), instead of eq.(3).

$$p_D(y = c_i | x) = \begin{cases} \exp(l_i) / \left(1 + \sum_{k=1}^{N} \exp(l_k)\right), & i = 1, 2, ..., N \\ 1 / \left(1 + \sum_{k=1}^{N} \exp(l_k)\right), & i = N+1 \end{cases} \tag{10}$$

The additional *N*+1 class is no longer required, and the discriminator turns into a standard classifier with *N* classes. The training procedure for the semi-GAN is same as that introduced in Section 3.2 except that the equation for calculating probability is changed from eq.(3) to eq.(10).

*3.4 Intuitive explanations about the principle of Semi-GAN using a simple example: how it utilizes unlabeled data to improve performance*

Fig. 5 shows a set of data samples which have two features (one feature corresponds to the x-coordinate and the other feature corresponds to the y-coordinate). The samples in the inner ring belongs to class 1 while the samples in the outer ring belongs to class 2. Consider the following scenario: although we have a large amount of data samples, the number of labeled samples is very limited: 3 samples in class 1 and 3 samples in class 2, as shown in Fig. 5. Assume that we do not know the labels of other data samples.

As for the supervised learning framework, only 6 labeled data samples can be used to train the classifier. All the unlabeled data samples are ignored. The decision boundary obtained by a supervised neural network (supervised NN) is plotted in Fig. 5 (the pink dotted line). We can see that only the labeled data samples can be well separated by this boundary. However, this decision boundary cannot separate "two rings".

The semi-GAN can utilize both unlabeled data and labeled data to train the classifier. From the unlabeled data samples, the semi-GAN will try to learn the distribution or pattern of the data. Specifically, in the example shown in Fig. 5, the semi-GAN will learn that the data form two rings. Then, with very few labeled data samples, the semi-GAN will learn that the data samples in the inner ring should be assigned to class 1 and the samples in the outer ring should be assigned to class 2. The green dotted line plotted in Fig. 5 is the decision boundary obtained by semi-GAN. Two rings are well separated by this boundary. Compared with the supervised NN, semi-GAN can significantly improve the classification performance by exploiting the information contained in unlabeled data.

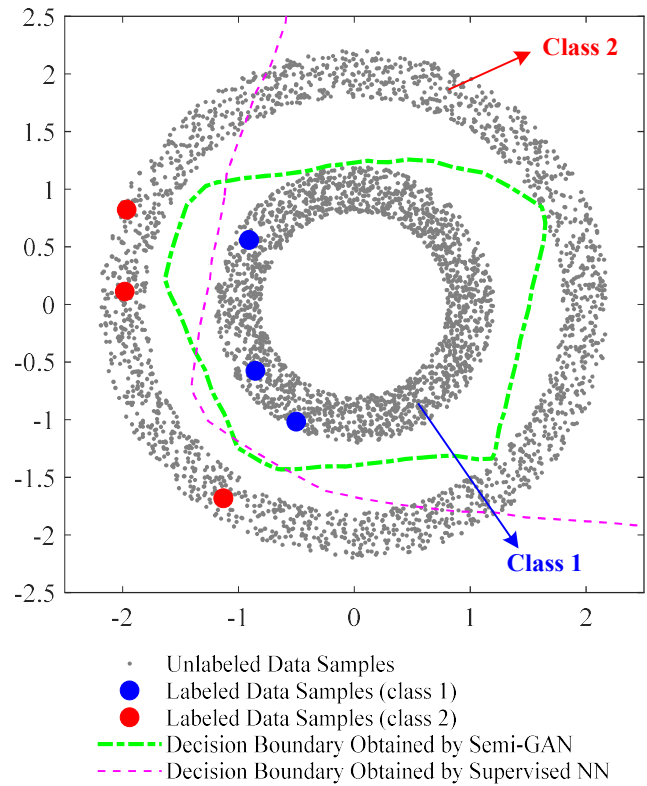

Fig. 5. Intuitive explanations about the principle of semi-GAN using a simple example

## 4. Proposed Fault Diagnosis Framework based on Semi-GAN

*4.1 Data preprocessing: remove the outliers and standardize*

The original training data should be preprocessed before presented to semi-GAN. This is because there exist some outlying data in the original dataset. These outlying data may be due to the erroneous sensor readings. These readings are obviously wrong and relatively infrequent. In this study, the following preprocessing steps are designed to remove the outliers in the original dataset:

1) Remove the data whose values do not lie within seven standard deviations of the mean. This step is to remove the outliers whose values are obviously deviated from the normal range.

2) Calculate the changing rates of the data. The changing rate of feature $x_1$ at time instant $t_1$ equals $|x_1(t_1+1) - x_1(t_1)|$. Then calculate the mean and the standard deviation of the changing rates. If the changing rates of a feature at current instant and next instant both lie outside seven standard deviations of the mean, the data point at current instant should be removed. This step is to remove the outliers whose values lie within the normal range (for the whole dataset) but differ significantly from adjacent points.

After removing the outliers, the data for each feature are standardized to the same interval [-1,1].

*4.2 Chiller fault diagnosis based on Semi-GAN*

The fault diagnosis problem can be regarded as a classification task. Different faults are assigned with different fault labels, as shown in Table 1. A classifier can be trained under the semi-GAN framework using the labeled data and the unlabeled data. The details of the semi-GAN training algorithm for chiller fault diagnosis is shown in Table 2. The trained discriminator network can be used as the fault classification model in the online testing. The hyperparameter settings of the semi-GAN for chiller FD are illustrated as follows:

The semi-GAN consists of two neural networks, a generator network, and a discriminator network. The input data has 61 features so the number of neurons in the input layer of the discriminator is set as 61. The number of features of the generated samples by the generator should be same as the samples in training set so the output layer of the generator has 61 neurons. In this study, the number of classes is 8, with 7 faulty classes and 1 normal class. The output layer of the discriminator has 8 neurons. Besides, we also set the size of the input layer of generator as 8.

As for the settings for the hidden layer size, the neural network with more hidden neurons means that this network can represent more complex patterns. Through many numerical tests, we found that the patterns of the chiller fault dataset are relatively simple, and it would be better to use a network with simpler structure as the discriminator. Besides, we also found that using the network with more complex structure as the generator can make the discriminator achieve better performance than using the simpler one. The reason may be that the generator with more complex structure can produce data samples closer to the real ones. And this forces the discriminator to improve its ability to discriminate the fake samples from the real samples (unlabeled data samples), which means that the distribution of the unlabeled data samples can be recognized more accurately by the discriminator. As a result, the performance of the discriminator can be improved. Therefore, in this study, we use a more complex network as the generator and a simpler network as the discriminator. The size of the generator network is chosen as 8-64-64-61 and the size of the discriminator network is 61-32-16-8.

The stochastic gradient descent algorithm can be used to train the semi-GAN. The literature [35] proposed an efficient variant of the gradient descent algorithm called Adam algorithm. The Adam algorithm can help to find better local minima and converge more quickly [35]. Thus, the Adam algorithm is adopted to train the semi-GAN in this study. During the training, the minibatch size $b_m$ is set as 128 and the number of training iterations is set as 100. Since the number of labeled samples is relatively small, all labeled samples are presented to semi-GAN as a whole batch. In this study, the training of the semi-GAN is implemented in Python 3.7 using Pytorch module. All the relevant codes have been uploaded in the supplementary materials.

Table 2: Algorithm of training semi-GAN

| |
|---|
| **Input:** labeled operating data of the chiller $x_l = \{x_l^{(1)}, x_l^{(2)}, ..., x_l^{(n_l)}\}$, labels of the labeled data $t = \{t^{(1)}, t^{(2)}, ..., t^{(n_l)}\}$, unlabeled operating data of the chiller $x_u = \{x_u^{(1)}, x_u^{(2)}, ..., x_u^{(n_u)}\}$. $x_u$ is divided into several minibatches with the batch size of $b_m$. |
| **for** number of training iterations **do** |
|     **for** each minibatch $\{x_u^{(1)}, x_u^{(2)}, ..., x_u^{(b_m)}\}$ **in** $x_u$ |
|       1: Sample minibatch of $b_m$ noisy samples $\{z^{(1)}, z^{(2)}, ..., z^{(b_m)}\}$ from the normal distribution $N(0,1)$. |
|       2: Obtain generated data samples $\{\tilde{x}^{(1)}, \tilde{x}^{(2)}, ..., \tilde{x}^{(b_m)}\}$, where $\tilde{x}^{(i)} = G(z^{(i)})$ |
|       3: Use unlabeled samples $\{x_u^{(1)}, x_u^{(2)}, ..., x_u^{(b_m)}\}$ to update discriminator parameters $\theta_d$ to minimize the loss function $L_{unlabeled}$ (given by eq.(5)). $\theta_d = \theta_d - \eta \nabla_{\theta_d} L_{unlabeled}$ |
|       4: Use generated data samples $\{\tilde{x}^{(1)}, \tilde{x}^{(2)}, ..., \tilde{x}^{(b_m)}\}$ to update discriminator parameters $\theta_d$ to minimize the loss function $L_{fake}$ (given by eq.(6)). $\theta_d = \theta_d - \eta \nabla_{\theta_d} L_{fake}$ |
|       5: Use labeled samples $x_l$ with labels $t$ to update discriminator parameters $\theta_d$ to minimize the loss function $L_{labeled}$ (given by eq.(4)). $\theta_d = \theta_d - \eta \nabla_{\theta_d} L_{labeled}$ |
|       6: Use generated data samples $\{\tilde{x}^{(1)}, \tilde{x}^{(2)}, ..., \tilde{x}^{(b_m)}\}$ to update generator parameters $\theta_g$ to minimize the function $L_3$ (given by eq.(8)). $\theta_g = \theta_g - \eta \nabla_{\theta_g} L_3$. The discriminator parameters $\theta_d$ are fixed when updating $\theta_g$. |
|     **end for** |
| **end for** |
| **Output:** discriminator parameters $\theta_d$. |

*4.3 Implementation demonstration of the proposed FD method in real chiller systems*

The design of the proposed fault diagnosis method does not involve the basic working principles of chiller systems. It is a data-driven method which learns how to recognize different faults from the training data. The proposed method can be easily applied to other types of chiller systems. The application of the proposed FD framework in real chiller systems is illustrated in Fig. 6.

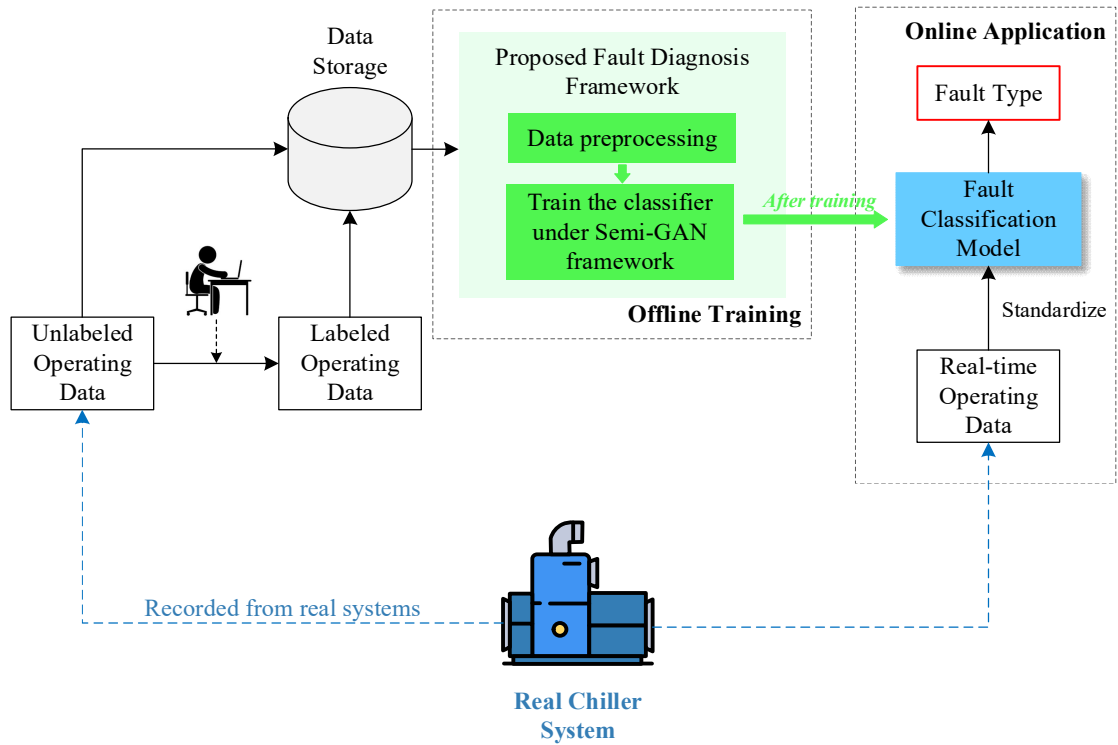

Fig. 6. Application of the proposed FD framework in real chiller systems

The implementation procedures of the proposed method can be summarized as follows:

(1) **Obtain labeled/unlabeled operating data in real chiller systems.** In practice, the data recorded in real chiller systems are mainly unlabeled. Label acquisition is laborious and costly. Only a small portion of data can be labeled manually by experienced experts or engineers according to prior knowledge or related field information.

(2) **Train the fault classifier using the labeled and unlabeled data.** Firstly, preprocess the original data obtained in previous step. Then, these data are used to train the fault classifier under semi-GAN framework. The training algorithm is given in Table 2. The trained discriminator network will be used as the fault classifier.

(3) **Online application.** An online fault diagnosis system can be designed based on the obtained fault classifier in step (2). The classifier takes in the real-time operating data measured in the real chiller systems and outputs the fault types. If a fault is diagnosed in the chiller system, the diagnosis system will report the corresponding fault alarm.

## 5. Validation Using Experimental Data

After removing the outlying data samples, the dataset has 162379 samples in total. This study involves four kinds of datasets: the unlabeled training set, labeled training set, validation set, and the testing set. These datasets can be formed by randomly choosing data samples from the original dataset. To mimic the practical scene that most of the historical data are actually unlabeled and only a small fraction of the data are labeled, the size of the unlabeled training set is set much larger than that of the labeled training set. Only very few data samples are chosen to form the labeled training set. The labels of the data samples in unlabeled training set are removed artificially and assumed to be unknown. The unlabeled training set together with the labeled dataset are used to train the semi-GAN. The obtained discriminator network (classifier) is used as the online fault classification model, which takes in the real-time operating data and outputs fault types.

*5.1 Fault diagnosis results of the proposed FD method compared with the supervised learning baseline*

To evaluate the performance of the fault diagnosis, a metric, diagnostic accuracy is introduced. The diagnostic accuracy is defined as the number of correctly diagnosed samples in the testing set to the total number of samples in the testing set. The performance of the proposed method is firstly compared with four supervised learning methods. The first one uses a one-layer neural network as the fault classifier. The number of the hidden neurons is chosen as 32 by the validation set. The second method uses a two-layer neural network as the fault classifier. The structure of this neural network is set the same as that of the discriminator network in semi-GAN. This setting is to avoid the effect of different network structures on the results, which can help sufficiently verify the merits of semi-GAN. The third and fourth methods are the recently proposed FD methods for chiller systems in existing literatures [23, 24], based on LS-SVM (least square support vector machine) and DBN (discrete Bayesian network), respectively.

The proposed FD method can utilize the information in the unlabeled data samples and obtain relatively high accuracy even using very limited number of labeled data samples. Testing experiments were conducted by varying the sizes of the labeled training set. The size of the labeled training set changes from 40 to 1600 while the unlabeled training set is fixed with 16000 samples. The four supervised learning method can only use the labeled set to train the fault classifier. To make fair comparison, the testing set is fixed (with 16000 samples) for different methods.

Since our datasets are randomly chosen from the original dataset, different random seeds will result in different results. Besides, different initialization of the training will also have some influence on the results. To reduce the variance and make our results more credible, the experiments in this study are conducted based on the following rules:
- In each scenario, the neural network (or other classifier) is trained ten times using ten different initial values. After the evaluation on testing set, ten diagnostic accuracy values can be obtained. The best one of these ten values is chosen as the final result for this scenario. This rule is to alleviate the effect of different initializations.
- We repeated the generation of different datasets (labeled, unlabeled, validation and testing sets) ten times and evaluate different methods by averaging the results over the ten experiments. This rule is to alleviate the effect of different division of datasets.

Table 3 shows the performances of different methods for various cases with different sizes of labeled training set, in terms of the average testing accuracy of ten experiments, with standard deviation in parentheses. As shown in Table 3, we can see that the diagnostic accuracy can be significantly improved especially for the case with less labeled training data. When there are only 40 labeled samples available, the proposed method is still able to diagnose the faults with an accuracy of 71.01% while the accuracies of other supervised methods are all below 45%. When there are 80 labeled samples, the proposed method can improve the diagnostic accuracy to 84%, while the supervised methods only reach the accuracy of 65% at most. By using the unlabeled data, the proposed method can significantly improve the diagnostic performance compared with the supervised methods, especially for the case with limited number of labeled samples.

Table 3: Testing accuracy using different methods with different sizes of labeled dataset

| Size of the labeled dataset | **Proposed Method** | NN (1 layer) | NN (2 layers) | LS-SVM [24] | DBN [23] |
|---|---|---|---|---|---|
| | Diagnostic Accuracy on Testing Set | | | | |
| 40 | **0.7101** (**0.0524**) | 0.4450 (0.0238) | 0.4412 (0.0332) | 0.3654 (0.0439) | 0.4278 (0.0241) |
| 80 | **0.8404** (**0.0329**) | 0.6479 (0.0323) | 0.6482 (0.0348) | 0.5960 (0.0359) | 0.6159 (0.0347) |
| 160 | **0.8914** (**0.0192**) | 0.7755 (0.0182) | 0.7861 (0.0168) | 0.7388 (0.0148) | 0.7536 (0.0194) |
| 320 | **0.9193** (**0.0065**) | 0.8512 (0.0102) | 0.8575 (0.0094) | 0.8224 (0.0116) | 0.8437 (0.0135) |
| 800 | **0.9441** (**0.0034**) | 0.9113 (0.0040) | 0.9091 (0.0038) | 0.9002 (0.0068) | 0.9061 (0.0036) |
| 1600 | **0.9579** (**0.0016**) | 0.9352 (0.0028) | 0.9336 (0.0034) | 0.9379 (0.0029) | 0.9370 (0.0030) |

Fig. 7 and Fig. 8 show the testing accuracy results for different faults using different methods when the size of the labeled dataset is fixed as 80. The accuracy increases when the severity level becomes higher. Compared with other methods, the performance improvement achieved by the proposed method for the fault with lower level of severity is more significant than that with higher level of severity. This means the faults can be accurately diagnosed in the incipient stage before they develop into the severe ones. The reason for this performance improvement is that semi-GAN can learn the information of the distribution of data from the large number of unlabeled samples. With the help of this information, it is easier for the classifier to identify the patterns of different faults. Therefore, the proposed FD method can diagnose the faults of low severity with much higher accuracy compared with other supervised FD methods.

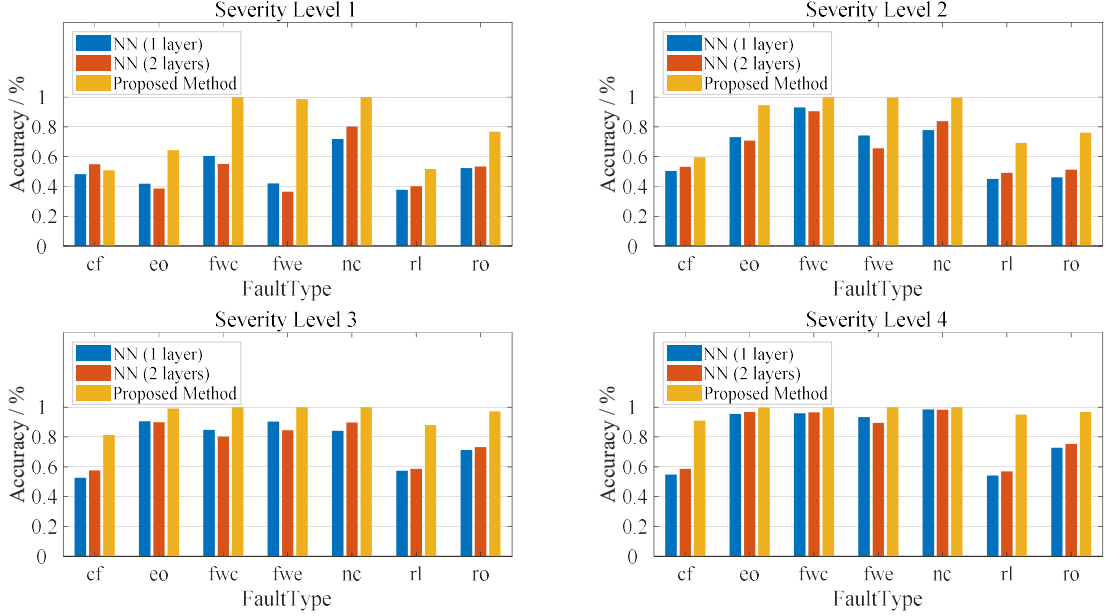

Fig. 7. Testing accuracy results for different methods (compared with NN-based methods) at 4 severity levels under the case with limited number of labeled training data samples (80) and large number of unlabeled training data samples (16000)

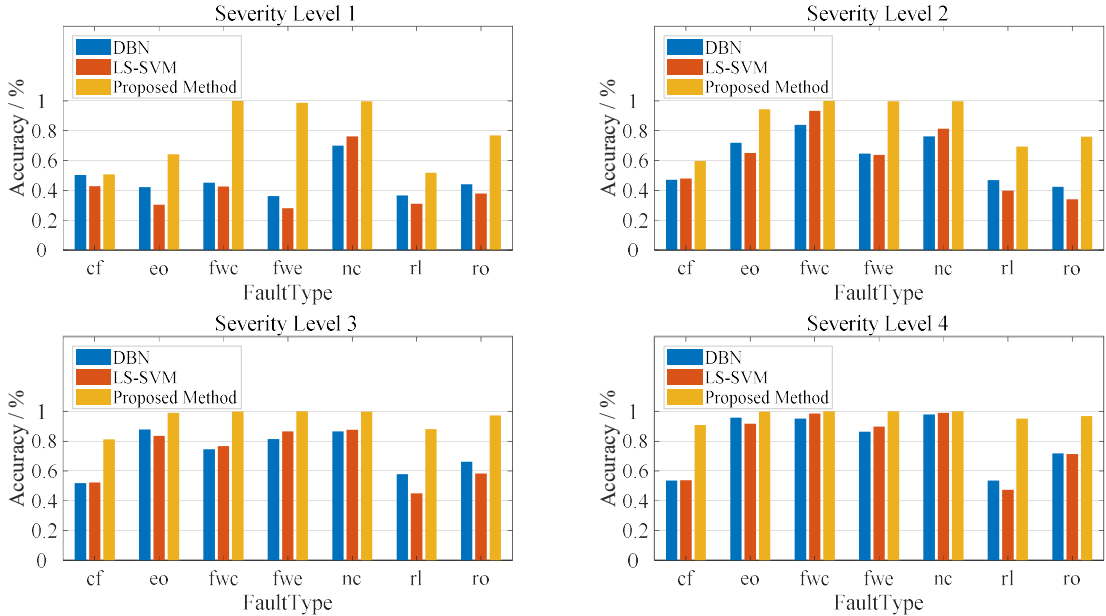

Fig. 8. Testing accuracy results for different methods (compared with LS-SVM [24] and DBN [23]) at 4 severity levels under the case with limited number of labeled training data samples (80) and large number of unlabeled training data samples (16000)

*5.2 Fault diagnosis results of the proposed FD method compared with the existing semi-supervised FD methods for chillers*

To further verify the effectiveness of the proposed method, the proposed FD method is compared with the existing semi-supervised FD method. There are several semi-supervised FD methods in the existing literatures. However, most methods mainly focused on the imbalanced data problem that fault samples are insufficient, instead of how to utilize the information in unlabeled data samples. Methods in [30] proposed a GAN-based semi-supervised framework to synthesize the faulty data. Method in [28] proposed a semi-supervised data-driven method to tackle the issue that data containing information about conditions related to faulty operation are usually unavailable. Only the semi-SVM-based FD method proposed in [29] provides a solution to utilizing the unlabeled samples. By using this method, the number of training samples can be enhanced by iteratively inserting highly confident unlabeled samples. After training, the obtained competing classifiers are evaluated on the testing set. The comparison results are listed in Table 4:

Table 4: Testing accuracy using different methods with different sizes of labeled dataset
(The results for the supervised learning are the best results among 4 supervised classifiers in Table 3)

| Size of the labeled dataset | **Proposed Method** | Semi-SVM [29] | Supervised Learning |
|---|---|---|---|
| | Diagnostic Accuracy on Testing Set | | |
| 40 | **0.7101** | 0.4696 | 0.4450 |
| | **(0.0524)** | (0.0364) | (0.0238) |
| 80 | **0.8404** | 0.6887 | 0.6482 |
| | **(0.0329)** | (0.0346) | (0.0348) |
| 160 | **0.8914** | 0.8020 | 0.7861 |
| | **(0.0192)** | (0.0156) | (0.0168) |
| 320 | **0.9193** | 0.8667 | 0.8575 |
| | **(0.0065)** | (0.0137) | (0.0094) |
| 800 | **0.9441** | 0.9254 | 0.9113 |
| | **(0.0034)** | (0.0050) | (0.0040) |
| 1600 | **0.9579** | 0.9512 | 0.9379 |
| | **(0.0016)** | (0.0035) | (0.0029) |

Compared with the results of the supervised learning, the semi-SVM-based FD method can slightly improve the diagnostic performance by 2%~5% while the proposed method can significantly improve the diagnostic performance by more than 20% when the number of labeled samples is less than 100. This demonstrates that the proposed FD method can utilize the unlabeled data samples in a more efficient manner.

*5.3 Effect of the unlabeled data size and the labeled data size*

In addition to different labeled data size, the effect of the unlabeled data size is also investigated in this study. Experiments were conducted by varying the sizes of unlabeled training set and labeled training set. The size of the unlabeled training set changes from 800 to 16000.

From Fig. 9, we can observe that all the accuracy values of the proposed method are above the supervised baseline. Increasing the size of unlabeled training sets can help to improve the diagnostic performance. The performance improvement with the number of unlabeled samples is more significant for the case when there are less labeled training samples. This is because for the case with more labeled training samples, the classifier can obtain enough ability to discriminate different faults from training with labeled samples. Besides, when there are very few labeled samples, the accuracy increases sharply in the beginning and then tends to be constant with the increase of unlabeled data samples. For the case with 40 labeled samples, using the proposed method with 800 unlabeled samples can increase the testing accuracy from 45% to 61%.

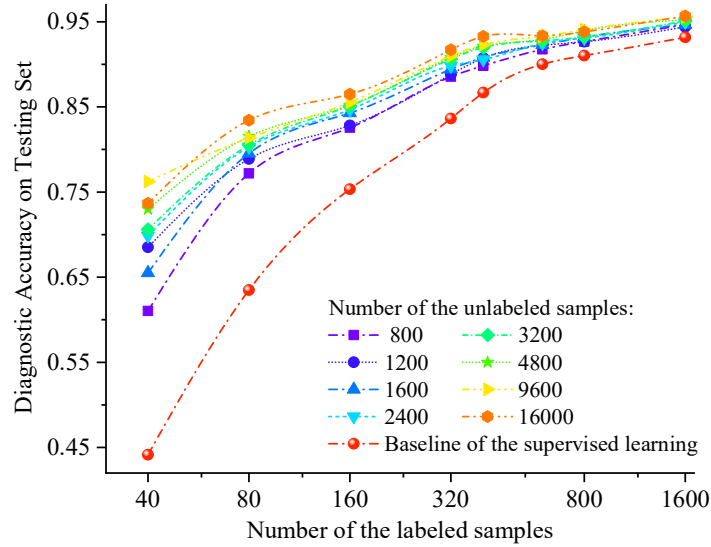

Fig. 9. Testing accuracy results of the proposed FD method with different number of unlabeled training samples and different number of labeled training samples

*5.4 Minimal required labeled data size for different desired diagnostic accuracy*

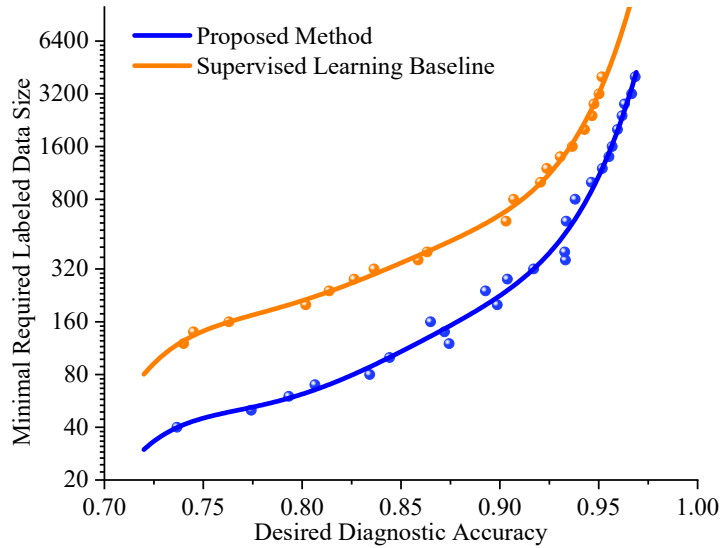

Fig. 10. Minimal required labeled data size for different desired diagnostic accuracy using the proposed FD method compared with the best supervised method (with 16000 unlabeled data samples). The y-axis is in log-scale.

Fig. 10 shows the minimal required labeled data size for different desired diagnostic accuracy using the proposed semi-GAN-based FD method compared with the supervised learning baseline. Two fitted lines are plotted through the data points. The supervised learning baseline is obtained based on the best results among 4 supervised methods in Table 3. As is seen in Fig. 10, to achieve higher diagnostic accuracy, more labeled training data are required. The minimal required labeled data size of the proposed FD method is much smaller than that of the supervised learning baseline. The semi-GAN can fully exploit the information contained in the large number of unlabeled data and thus reduce the requirement for the labeled data. Fig. 11 shows the reduction in labeled data size using the proposed semi-GAN-based FD method compared with the supervised method. As seen, the reduction in labeled data size increases with the unlabeled data size. With the increasing number of unlabeled data samples, more information can be utilized. The proposed method can achieve the desired diagnostic accuracy with less labeled data samples. When there are plenty of unlabeled samples (e.g., more than 4800), the minimal required number of labeled samples can be reduced by about 60% with the proposed method compared with the supervised learning baseline. The proposed method demonstrates its superiority especially in the scenario with limited number of labeled samples and plenty of unlabeled samples.

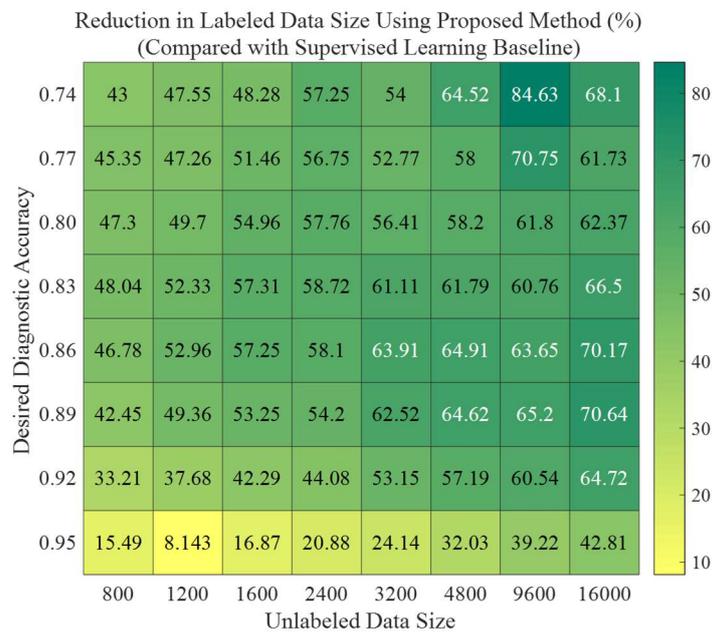

Fig. 11. The reduction in labeled data size using the proposed FD method compared with the supervised baseline

## 6. Conclusion

This paper proposes a novel semi-supervised data-driven fault diagnosis method for the chiller systems based on the semi-generative adversarial network. Both the labeled data and the unlabeled data are incorporated into the learning process. The proposed method can learn the information of data distribution from the unlabeled data and thus make effective use of the unlabeled data to improve the fault diagnosis performance. The experimental data from ASHRAE Project 1043-rp are used to evaluate the performance of the proposed method. The main findings of this paper can be summarized as follows:

- The proposed method can effectively utilize the unlabeled data to improve diagnostic performance. When the number of labeled samples is as small as 80, the proposed method can achieve higher diagnostic accuracy of 84% while the supervised baseline methods only reach the accuracy of 65% at most.
- The performance improvement achieved by the proposed method for the fault with lower level of severity is more significant than that with higher level of severity. This means that the faults can be accurately diagnosed in the incipient stage before they develop into the severe ones.
- Increasing the size of unlabeled training sets can help to improve the diagnostic performance. The performance improvement with the number of unlabeled samples is more significant for the case when there are less labeled training samples.
- The proposed method can significantly reduce the required labeled data size. Much less labeled data are required for obtaining the desired diagnostic accuracy. This is because the semi-generative adversarial network can fully exploit the information contained in unlabeled data. The minimal required number of labeled samples can be reduced by about 60% with the proposed method when there are enough unlabeled samples.

The merits of the proposed method make it easily to apply in the practical use, especially in the practical scenario with limited number of labeled data and plenty of unlabeled data. The limitation of this study is that the proposed method cannot diagnose the unknown faults in the chiller system. The unknown faults are the faults which are not included in training datasets. Besides, this study cannot diagnose two faults which occur simultaneously. The future work of this study includes:

- Improve the current fault diagnosis framework to incorporate the diagnosis of unknown faults of the chiller system. It is unrealistic to simulate every possible fault in the experiment so unknown faults may occur in real systems. New methods should be proposed to accurately diagnose the unknown fault.
- Develop new fault diagnosis methods which can diagnose concurrent faults in the chiller system. The existing studies focus on diagnosing a single fault at one time. In practice, it is possible that two faults can occur simultaneously. The characteristics of multiple faults may be more complex and should not be the simple superposition of the characteristics of each fault alone.